\documentclass[runningheads]{llncs}

\usepackage{amsthm}
\usepackage{graphicx}
\usepackage{subfig}
\usepackage{comment}
\usepackage{amsmath,amssymb} 
\usepackage{color}
\usepackage{hyperref}
\usepackage{bbm}
\usepackage{mathabx}
\usepackage{multirow}
\usepackage{caption}
\usepackage{bbm}
\usepackage[ruled,vlined,linesnumbered]{algorithm2e}
\newcommand{\norm}[1]{\left\lVert#1\right\rVert}

\begin{document}
\pagestyle{headings}
\mainmatter
\def\ECCVSubNumber{5186}  

\title{Exploiting Non-Linear Redundancy for Neural Model Compression} 

\titlerunning{ECCV-20 submission ID \ECCVSubNumber} 
\authorrunning{ECCV-20 submission ID \ECCVSubNumber} 
\author{Anonymous ECCV submission}
\institute{Paper ID \ECCVSubNumber}

\titlerunning{Abbreviated paper title}
%
\author{Muhammad A. Shah \and
Raphael Olivier \and
Bhiksha Raj}
\authorrunning{Shah et al.}
%
\institute{Carnegie Mellon University, \\Pittsburgh, PA, USA \\\email{\{mshah1, rolivier, bhiksha\}@cs.cmu.edu}}
\maketitle

\begin{abstract}
Deploying deep learning models, comprising of non-linear combination of millions, even billions, of parameters is challenging given the memory, power and compute constraints of the real world. This situation has led to research into model compression techniques most of which rely on suboptimal heuristics and do not consider the parameter redundancies due to linear dependence between neuron activations in overparametrized networks. In this paper, we propose a novel model compression approach based on exploitation of linear dependence, that compresses networks by elimination of entire neurons and redistribution of their activations over other neurons in a manner that is provably lossless while training. We combine this approach with an annealing algorithm that may be applied during training, or even on a trained model, and demonstrate, using popular datasets, that our method results in a reduction of up to 99\% in overall network size with small loss in performance. Furthermore, we provide theoretical results showing that in overparametrized, locally linear (ReLU) neural networks where redundant features exist, and with correct hyperparameter selection, our method is indeed able to capture and suppress those dependencies.
\end{abstract}

\section{Introduction}
Modern Deep Neural Networks (DNN) have pushed the state-of-the-art in many computer vision tasks including image recognition, object detection, etc. Underlying the success of DNNs, are the millions of constituent parameters and their non-linear combinations that allow DNNs to accurately model a wide variety functions over their input features. However, running inference over these massive models imposes exorbitant memory and computational costs that make deploying them at scale in the real-world with more stringent latency, compute and energy constraints, a challenging problem. 

This problem is exacerbated if later models build on existing models, which are themselves very large. Often, the intermediate outputs of a pretrained image recognition model are used to encode visual information for another downstream task, such as visual question answering, image retrieval, etc. Since these intermediate outputs can be very high-dimensional, using them as features can introduce a large number of parameters to downstream models. Therefore, methods for compressing the neural models
can have far reaching effects vis-a-vis their deployability and scalability.

As deep learning models make their way from research labs to real world environments, the task of making them more resource efficient has received a great deal of attention. This has given rise to a large body of work focused on compressing and/or accelerating DNNs. One of the most common techniques for compressing neural models is parameter pruning, i.e. pruning away model parameters or neurons based on some metric. Such methods have many limitations ; one of them is that the heuristic used attempts to identify "weak" elements (with small magnitude or derivative, for instance), but parameters can be unnecessary in more ways than being small. Consider two neurons taken individually that both have large amplitude, but happen to yield identical outputs ; one of them can be pruned without loss of information, but most current pruning methods could not figure it out.

In this paper we propose a novel neural model compression technique that exploits the dependencies in the non-linear activations of the units in each layer to identify redundant units and prune them. Our technique is based on the observation model optimization can potentially converge to a point at which the outputs of several units in one layer become highly correlated, even linearly dependent, with each other, and thus, by removing one of them and adjusting the outgoing weights of the other units we can obtain a smaller model with identical outputs. 

We identify redundant units by measuring the degree to which they can be predicted as a linear combination of other units in the same layer. Specifically, we learn a transformation matrix, $A$, that best approximates an identity mapping for the activations of the units in this layer, while constraining the diagonal of $A$ to be zero. We select the units with the lowest prediction error to remove, and adjust the outgoing weights of the remaining units using the values in the corresponding columns $A$ such that input to the next layer remains the same (or is minimally perturbed). Once we have removed all the predictable units, removing any additional units will cause a reduction in model performance. We then fine-tune the model to recover the lost accuracy. In order to facilitate tuning, we use distillation \cite{hinton2015distilling} to bring the compressed model's output distribution close to the uncompressed model's output distribution raised to a high temperature.

We demonstrate the efficacy of our technique on two popular image recognition models, AlexNet\cite{alexnet} and VGG\cite{vgg}, and three popular benchmark datasets, CIFAR-10\cite{cifar10} and Caltech-256\cite{caltech256}. We demonstrate, theoretically and empirically, that under our proposed weight readjustment scheme, the inputs to the subsequent layer are only minimally perturbed while redundant units are present. Our technique can reduce the parameters of VGG and AlexNet by more than 99\% on CIFAR10 and by more than 80\% on Caltech-256. Finally, we inspect the intermediate representations of the compressed models and show that the data remains cleanly separable post-compression, which suggests that the intermediate representations continue to capture rich information about the data which may be useful for transfer learning tasks.

\section{Related Work}
Most existing techniques for reducing the size of neural models can be grouped into three high-level categories, namely low-rank factorization, knowledge distillation and, parameter pruning. We argue that all of them have significant shortcomings. There are also methods that reduce computation while not really affecting the number of parameters in the network,  such as quantization \cite{Han2016}\cite{oland2015reducing}; however these are somewhat orthogonal to the scope of this paper.

\subsection{Pruning}
A common pruning approach to model compression attempts to eliminate individual parameters in the network \cite{touretzky1996advances,xing2020probabilistic,LIU2020Dynamic}. Many of them do so by enforcing a sparsity constraint, such as $L1$ regularization, to push some parameters to $0$, or $L2$ regularization to simply keep weights small and then prune the small ones \cite{han15}. Those methods can achieve reasonable performance (up to 35x compression in \cite{Han2016}). One of theses methods' main limitations however, is that their outputs take the form of sparse weight matrices, and benefiting of those in terms of computation time is not always easy in practice.

A different family of methods overcome that shortcoming by pruning out entire neurons and/or convolution filters \cite{Li17,liu2017learning,zhuang18,mussay2020dataindependent,peng2019collaborative}. Those methods identify weak neurons using heuristics such as activation value \cite{Hu16} or absolute sum of weights \cite{Li17}. Fine-tuning may be used afterwards \cite{liu2017learning}.

Both of these methods treat each unit independently, in that they prune the unit if that unit has little effect on the downstream computations. However, these techniques would not prune a unit whose output is significantly impacts downstream computation but is largely predicted as a linear combination of the outputs of the other units in the same layer. Such units can be pruned, and their weights distributed, to achieve lossless compression.

\subsection{Matrix Factorization}
Factorization-based approaches \cite{Denton14,lu2017fully,yang2015deep} factor the weight matrices of the neural network into multiple low-rank components which can be then used to approximate the output of the original weight matrix. Those techniques are seemingly more similar to our approach : low-rank matrices eliminate some redundancies by projecting on a subspace of smaller dimension. 

The key difference is that those methods work at the weight matrix level, while we find redundancies within post-activation representations. The non-linearity of those activations is likely to create new redundancies, that escape factorization methods but that our approach can capture. However, the case of a linear activation is similar to a factorization process, and we will to some extent use it in \ref{sec:math} to better understand the process of weight readjustment and how it affects the error surface.

\subsection{Distillation}
Model compression using knowledge distillation \cite{hinton2015distilling} involves training a more compact student model to predict the outputs of a larger teacher model, instead of the original targets \cite{luo2016face,belagiannis2018adversarial}. Distillation, or annealing, provides an interesting angle to the compression problem, but in itself suffers from several shortcomings. First, it requires the student model to be predefined, which is a rather tedious task. By default this model would need to be trained from scratch. Finally, this process relies solely on the extra supervision of the teacher model to overcome the challenges of training a less overparametrized model, with complex error surfaces ; this seems sub-optimal. 

On the contrary, distillation can become very useful as a complementary compression tool. Assuming a compression method induces some drop in performance in the smaller model, a short amount of fine-tuning may boost its performance, and using knowledge distillation from the original model at that step can speed up that process. We make use of distillation in such a manner, in an iterative fashion, and discuss it in \ref{sec : annealing}

\section{Lossless redundancy elimination : formalism and justification}
\label{sec:math}
\subsection{Notations and task definition}
Throughout the paper, we consider a feed-forward neural network
$F = \phi_N\circ L_N\circ\phi_{N-1}\circ L_{N-1}\circ ...\circ\phi_1\circ L_1$ where $L_k$ is the k-th dense or convolutional (which is dense with shared parameters) layer and $\phi_k$ is the following activation, in the largest sense. For example $\phi_k$ may involve a pooling, or a softmax operator in the case of $\phi_N$. The weight matrix of $L_k$ is $W_k$, of size $(n_k^o,n_{k+1}^{i^2})$. Depending on the nature of $\phi_k$, $n_k^i$ and $n_k^o$ may be different ; however, to alleviate notations later on, we will consider that $n_k^i=n_k^o =n_k$ which does not induce a loss of generality in our algorithms.

For a given input vector $X$, sampled from data distribution $\mathcal{D}$ we define the intermediate representations $Z_k$ (activations) and $Y_k$ (pre-activation), such that
$$Z_0=X\;\; ; \;\;Y_k = W_k.Z_{k-1}\;\; ; \;\;Z_k = \phi_k(Y_k)$$

Our goal is to eliminate redundancies within the activations of a given layer. To do so, we consider for each activation $Z_k[i]$ the task of predicting it as a linear combination of the neighbouring activations $Z_k[j],j\neq i$. Solving that task, evaluated with the $L_2$ norm, amounts to solving the following problem :
$$\min_{A_k\in \mathcal{M}_n(\mathbf{R})} \mathbf{E}_{x\sim D}[\norm{Z_k-A_kZ_k}_2^2] \;\;\;s.t.\; diag(A_k)=0$$

\subsection{Expression of the $A_k$ matrix}

Let us find the expression of redundancy matrix $A_k$.

We start by simplifying the problem's formulation. Rewriting the objective $\min_{A_k\in \mathcal{M}_n(\mathbf{R})} \sum_{i=1}^{n_k}\mathbf{E}_{x\sim D}[(Z_k[i]-A_k[i].Z_k)^2]$ ($M[i]$ being the ith row vector in matrix $M$), elements of $A$ of different rows are clearly uncoupled both in the objective and the constraint. We can therefore solve that problem row-wise. Writing $U = I_{n_k}-A_k$, we must solve $n_k$ problems $P_l$ of the form $$\min_{u\in\mathbf{R}^n_k}\mathbf{E}_{x\sim D}[(u^TZ_k)^2]\;\;\;s.t\; u_l=1$$.

Define 
$$g(u)=\mathbf{E}_{x\sim D}[(u^TZ_k)^2]
    = \mathbf{E}_{x\sim D}[(u^TZ_k).(u^TZ_k)^T]
    = \mathbf{E}_{x\sim D}[u^TZ_kZ_k^Tu]
    = u^TSu$$
where $S = \mathbf{E}_{x\sim D}[Z_kZ_k^T]$ is $Z_k$'s correlation matrix, which is positive semidefinite. Let's consider only the non-degenerate case where it is positive definite. $g_l$ is convex (of hessian $\frac{1}{2}S$, so the zeros of the gradients indicate exactly its minima). 

Specifically, within the hyperplane of admissible points $H={u\in \mathbf{R}^n_k,u_l=1}$, $g_l$ is minimal if and only if : 
$\nabla_ug = 2Su \in H^\bot = \mathbf{R}e_l$, where $e_l$ is the lth vector of the canonical base. Or in other words, $u$ is a minimum iff it is a multiple of $S^{-1}e_i = S^{-1}[l]^T$, the lth column of $S^{-1}$. Therefore the minimum is $u^* = \frac{1}{S^{-1}[l][l]}S^{-1}[l]^T$.

In full matrix form, we conclude that the solution $A_k$ to the equation is
$A_k = S^{-1}D$, with $S = \mathbf{E}_{x\sim D}[Z_kZ_k^T]$ and $D = diag(S^{ -1})^{-1}$

In practice, due to the presence of a matrix inversion, it is simpler and faster to obtain $A_k$ using gradient descent, given inputs sampled in the training set.

\subsection{Weight readjustment} 
\label{sec : readjust}
The previous weight matrix, and the residual error  provides information regarding which activation is most predictable and should be removed. We then wish to adjust the remaining weight matrix of the following layer to account for this redundancy elimination. Assume only one activation $l$ was removed. We consider the compressed vectors $Z_k^l$ (of size $n_k-1$) where activation l was removed. We can infer a transformation matrix $T_k^l$ from $Z_k^l$ to $Z_k$, with minimal error, using the lth row of $A_k$ : $Z_k \approx T_k^lZ_k^l$, where :
$$\forall{i<l}, T_k^l[i][j] = \delta_{ij}\;\;\forall{i>l}, T_k^l[i][j] = \delta_{i(j+1)}$$
$$\forall{j<l}, T_k^l[l][j] = A_k[l][j]\;\;\forall{j\geq l}, T_k^l[l][j] = A_k[l][j+1]$$

i.e. $ T_k^l$ is the identity matrix $I_{n_k-1}$ where $A_k[l]$ has been inserted as the lth row, minus the $0$ coefficient $A_k[l][l]$.

Therefore, a natural adjustment for the weight matrix is  $W_{k+1}^l=W_{k+1}T_k^l$, such that $W_{k+1}^lZ_k^l \approx W_{k+1}Z_k$.

If we remove more than one coefficient at once, the expression of the transformation matrix is less straightforward. Assume for example that activations $l$ and $j$ are removed ; Obtaining the approximate expression of $Z_k[l]$ obtained from $A_k$, using only remaining activations, leads to the following derivations :

$$Z_k[l] = \sum_{i=1}^{n_k}A_k[l,i]Z_k[i] = \sum_{i=1\\i\neq l,m}^{n_k}A_k[l,i]Z_k[i] + A_k[l,j]Z_k[j]$$
$$Z_k[l] =A_k[l,j]A_k[j,l]Z_k[l] \sum_{i=1\\i\neq l,j}^{n_k}A_k[l,i]Z_k[i]+ A_k[l,j]A_k[j,i]Z_k[i]$$
$$Z_k[l] =\frac{\sum_{i=1\\i\neq l,j}^{n_k}A_k[l,i]Z_k[i]+ A_k[l,j]A_k[j,i]Z_k[i]}{1-A_k[l,j]A_k[j,l]}$$

More generally, assume a set $J = \{j_1<...<j_m\} \subset [1..n_k]$ of activations is eliminated. We note its complementary set $H=[1..n_k]\setminus J = \{h_1<...<h_{n_k-m}\}$. We define $A_k^{J+}$, the $(m,m)$ matrix defined by $A_k^{J+}[i][p]=A_k[j_i][j_p]$, and $A_k^{J-}$, the $(m,n_k-m)$ matrix such that $A_k^{J-}[i][p]=A_k[j_i][h_p]$. Then we can write, where $Z_k^{J}$ contains only the activations in $J$ and $Z_k^{H}$ only those not in $J$: $$Z_k^{J} = A_k^{J+}Z_k^{J}+A_k^{J-}Z_k^{H}$$ i.e. $Z_k^{J} = (I_m-A_k^{J+})^{-1}A_k^{J-}Z_k^{H} = U_k^JZ_k^{H}$
provided that $(I_m-A_k^{J+})$ is invertible. From there we can easily obtain the equivalent of the transformation matrix in the single activation case :$Z = T_k^JZ_k^{H}$, where $T_k^J$ is $U_k^J$ completed with ones on the diagonal for the additional rows.
\subsection{Stability results on compression}
In practice, we wish to compress a network at some point during training, followed by  further fine-tuning. We may expect that if we are close enough to a good local minimum in a convex region of the error surface, we may capture that minimum's redundancies and eliminate them. One question however arises : after compression, will we still be close enough to the minimum that the fine-tuning will converge, or may it diverge? The following stability result gives partial answers to that question.

\theoremstyle{definition}
\begin{theorem}
Let the kth activation in the neural network be linear, so that we can write $F(X) = f_\theta(W_{k+1}.W_k.g_\phi(X))$ for $X$ in the training set $\mathcal{X}$. Let $p^* = (\theta^*,\phi^*,W_{k+1}^*,W_k^*)$ a local minimum for the loss function $L$ in the parameter space $\mathcal{P}$, and assume an exact redundancy of the final activation of the kth layer : $W_{k+1}^{*}.W_k^{*}=W_{k+1}^{*\prime}.W_k^{*\prime}$, where $W_k^{*\prime}$ is $W_k^{*}$ minus its final row, and $W_{k+1}^{*\prime}$ is the contracted weight matrix as computed in \ref{sec : readjust} with matrix $A_k$. The compression-readjustment operation projects $p*$ onto $p*' = (\theta^*,\phi^*,W_{k+1}^{*\prime},W_k^{*\prime})$

Assume there is a ${L_2}$-ball $B\subset \mathcal{P}$ of radius $R$ centered $p*$, on which $L$ is convex. There is an ellipsoid $E$ centered on $p*$ of equation $$\norm{\theta-\theta^*}_2^2+\norm{\phi-\phi^*}_2^2+\norm{W_{k+1}-W_{k+1}^*}_2^2+\sum_{i=1}^{n_k-1}\sum_{p=1}^{n_k}(1+A_k[n_k][i]^2)V[i][p]^2 \leq R$$ onto a convex region of the compressed parameter space.
\end{theorem}

We delay the full proof to an appendix. We can however discuss the implications of the theorem. We note first that the theorem by itself assumes linear activations. Assume our training brought us near a (global or local) minimum displaying some redundancies that we manage to eliminate. We would like regularities of the error space in that region, such as convexity, to be preserved after compression. That the point before compression was in a convex region around the minimum is not enough to have convexity in the compressed space; however, our theorem shows that there is a slightly different region, determined by the radius of the convex region around the minimum, that does ensure post-compression convexity. That region is obtained by flattening the ball onto the subspace of $W_{k+1}$. Besides, the subspace corresponding to the compressed coefficient can be ignored.

While the above theorem applies only to the linear activation case, we argue that the results extends naturally to locally linear or nearly linear activations. Consider for example a ReLU activation: around any parameter point, there is a ball on which the network is identical to one with linear activation; we can apply theorem 1 on that restricted area of the space.

\subsection{An empirical justification for readjustment}

\begin{figure}[t!]
    \centering
    \subfloat[]{\includegraphics[scale=.35]{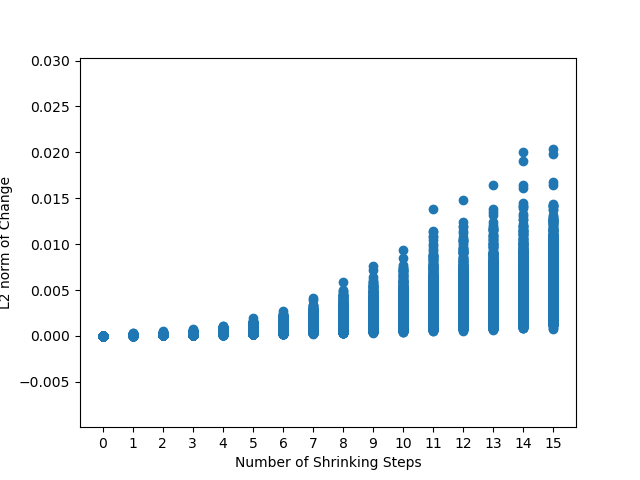}}
    \subfloat[]{\includegraphics[scale=.35]{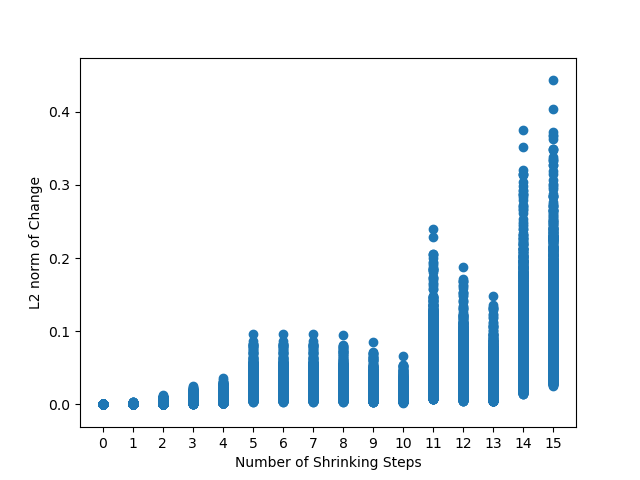}}
    \caption{\label{fig : norm}\small The norm of the change in the inputs to layer 17 of AlexNet after shrinking layer 16 down to almost 1\% of its original size (a0 with  and (b) without weight readjustment. Every step on the x-axis represents a shrinking step in which the layer is shrunk by 25\%.}
    \vspace{-10pt}
    \label{fig:my_label}
\end{figure}

The proposed pruning approach is based on the hypothesis that elimination of linearly dependent (or almost dependent) neurons (or filters) within any layer with appropriate weight adjustment will result in minimal or even zero changes to the representations seen by {\em subsequent} layers in the network.  In other words, compressing the  $k$th layer of a network as proposed should not significantly change the pre-activation values observed by layer $k+1$.  We evaluate this hypothesis in this section.

In Fig. \ref{fig : norm}, we plot pre-activation norm differences on an AlexNet network for an intermediate convolutional layer, computed on a random input sample, after compressing the previous layer. The norm difference is computed as 
$\frac{\norm{Z_{17}^{(i)} - Z_{17}^{(0)}}_2}{\norm{Z_{17}^{(0)}}_2}$, where $Z_{17}^{(i)}$ represents
the activations after the $i^{th}$ shrinking step. We compare the results obtained from just trimming dependent neurons without subsequent adjustment of weights, to those obtained after weight readjustment. As expected, trimming neurons modifies $Z$, but subsequent weight readjustment largely eliminates the changes from trimming -- after 15 compression steps we have only a $2\%$ norm change, confirming the intuition behind our method.

\section{When Activations Are Not Dependent}
The above analysis shows that if there is perfect linear dependence in the neuron activations, i.e. $\norm{Z_k-A_kZ_k}_2^2=0$, then we can achieve lossless compression, however, in many cases this condition may not hold. In such situations, the parameters of the pruned model, even after readjustment, may end up in a suboptimal region of the loss surface. This is because readjustment weights in $A$ are imperfect and error prone, and therefore will move the model parameters to a different, potentially suboptimal, point on the error surface. Since, reducing the size of the model makes the error surface less smooth \cite{allen2018convergence}, even if the operating point of the smaller model is close to the operating point of the larger model, it may have a much higher loss. To keep the model parameters from deviating too far from the optima during compression we employ a modified version of Annealed Model Contraction (AMC) \cite{shah2019annealing}, which attempts to keep the model in the optimal region by cycles between pruning and fine-tuning phases. Below we provide a description of AMC, and our modifications to it.


\subsection{Annealed Model Contraction}\label{sec : annealing}
AMC is an iterative method that greedily prunes the model layer-by-layer. As formulated in \cite{shah2019annealing}, AMC starts from the first (or the last) layer of the model and proceeds to maximally shrink the current layer before moving on to the next. While compressing a layer, AMC alternates between pruning and fine-tuning. Pruning is performed by reinitializing the layer with $\gamma\%$ fewer neurons/filters and the whole network is then fine-tuned end-to-end. During fine-tuning, knowledge distillation \cite{hinton2015distilling} is used to facilitate training and the following loss is minimized
\begin{equation}
    \label{eq:amc-loss}
    \mathcal{L} = (1-\lambda)H\left(\text{softmax}\left(\frac{\mathbf{z}}{T}\right),\text{softmax}\left(\frac{\mathbf{v}}{T}\right)\right) + \lambda H\left(\mathbf{y_{true}}, \text{softmax}\left(\frac{\mathbf{v}}{1}\right)\right)
\end{equation}
Where $\mathbf{z}$ and $\mathbf{v}$ are the logits returned by the teacher and student models, respectively, $T$ is a hyperparameter referred to as the temperature of the distribution, and $\lambda$ controls the contribution of the loss against the target label to the the overall loss.

AMC continues to prune a layer as long as the pruned model's accuracy remains within a threshold, $\epsilon$, of the uncompressed model's accuracy. Once the current layer can not be pruned any further, AMC proceeds to shrink the next layer in the model. AMC can be applied to both, dense and convolutional layers. In the case of the former, it prunes neurons while in the latter it prunes convolutional filters.

\subsection{Annealed Model Contraction with Lossless Redundancy Elimination}
\begin{algorithm}
\SetAlgoLined
 $\text{RemoveAndAdjust}(A, W, j)$: adjust the weight matrix after the removal of the $j^{th}$ neuron from the previous layer using the method in \ref{sec : readjust}\\
 \SetKwFunction{FLREShrink}{LREShrink}
 \SetKwProg{Fn}{Function}{:}{}
 \Fn{\FLREShrink{$F$, $l$, $\gamma$}}{
     $Z \leftarrow F_{1:l}(\mathcal{X})$\tcp{compute the activations of the $l^{th}$ layer.}
     $A \leftarrow \min_A \norm{ZA - Z}^2 \text{s.t } \text{diag}(A)=\mathbf{0}$\\
     $\mathcal{E} \leftarrow \arg\text{sort}(\norm{ZA - Z}^2)[:\lfloor\gamma * \text{sizeof}(F[l])\rfloor]$\\
     $\Bar{W}^{(l+1)}\leftarrow [W^{(l+1)}; b^{(l+1)}]$\tcp{Concatenate the weights and bias.}
     \For{$j\in \mathcal{E}$}{
        $W^{(l)} \leftarrow W^{(l)}_{-j.}$ \tcp{drop the $j^{th}$ row of $W^{(l)}$}
        $W^{(l+1)} \leftarrow \text{RemoveAndAdjust}(A, \Bar{W}^{(l+1)}, j)$\\
     }
  }
 $Acc\leftarrow \text{evaluate}(F_t)$\\

 $F_s'[i_B]\leftarrow\text{LREShrink}(F_s, i_B, \gamma)$\\
 $Acc'\leftarrow \text{evaluate}(F_s')$\\
 \While{$Acc-Acc'\leq\epsilon$}{
    $F_s\leftarrow F_s'$\\
    $F_s'[i_B]\leftarrow\text{LREShrink}(F_s',i_B, \gamma)$\\
    $Acc'\leftarrow \text{evaluate}(F_s')$\\
    \If{$Acc-Acc'>\epsilon$ }{
        $F_s'\leftarrow \text{distill}(F_s')$\\
        $Acc'\leftarrow \text{evaluate}(F_s')$\\
    }
 }
\caption{\label{alg:LRE-AMC}LRE-AMC Algorithm}
\end{algorithm}

While effective, AMC has the shortcoming that it takes an ad-hoc approach to parameter pruning. AMC removes neurons from a layer by \textit{reinitializing} the layer with fewer neurons. The new initialization is random and therefore can land the model arbitrarily far away from the optimal point. On the other hand, the Lossless Redundancy Elimination (LRE) formalism presented in Section \ref{sec:math} provides a method of setting the parameters of the pruned layer that guarantees (under some constraints) that the model remains near the optimal operating point. However, LRE only considers the activations and weights between two layer, and thus does not account for the effects of pruning on the operating point of the whole model. Therefore, we propose to combine LRE and AMC in a novel model compression algorithm, which we call LRE-AMC, that compensates for the inadequacies of both LRE and AMC.

LRE-AMC (Algorithm \ref{alg:LRE-AMC}) differs from vanilla AMC in two significant ways. \textit{First}, instead of pruning neurons by reinitializing the layer, LRE-AMC uses the LRE formalism to select neurons/filters (in the following we will use the term \textit{units} to refer to neurons and filters) to prune away based on the degree of linear dependency between their activations. Thus, LRE-AMC retains units that have linearly independent activation and thus have learned to encode unique aspects of the data, whereas these units have to be relearned under AMC. \textit{Second}, LRE-AMC breaks the pruning process into two phases. In the first phase LRE is used to remove the selected units one-by-one and adjust the weight matrix such that the outputs of the layer are minimally perturbed. After each pruning stage we measure the performance of the model on a held-out set and continue pruning without fine-tuning until the performance of the model remains within a threshold, $\epsilon$, of the original. When the performance drops below the threshold $\epsilon$, we start phase two in which we use distillation to fine-tune the model to bring the model's performance to within $\epsilon$ of the pre-compression performance.

\section{Evaluation}
\subsection{Datasets}
\label{sec:dataset}
We evaluate our proposed method three datasets of real-world images, namely CIFAR-10 and Caltech-256. CIFAR10 contains 50,000 training images and 10,000 testing images. Each images has a size of $32\times32$ pixels and is assigned one out of 10 class labels. Caltech-256 contains 30,607 real-world images, of different sizes, spanning 257 classes. Following the protocol from \cite{vgg}, we construct a balanced training set for Caltech 256 with 60 images per class. For both, Caltech256 and CIFAR10, we used 20\% of the training images for validation during training. We apply data augmentation to increase the size of the training data and improve generalization. Specifically, we augment CIFAR10 with random affine transformations, horizontal flips and grayscaling. Meanwhile, we augment Caltech-256 by taking a random $256\times256$ crop at a random scale between 0.8 and 1.0, and applying rotation, color jitter and horizontal flipping before resizing the image to $224\times224$. The pixel values of images from both datasets are normalized by mean [0.485, 0.456, 0.406] and standard deviation [0.229, 0.224, 0.225].

\subsection{Experimental Setup}
We implemented AMC and LRE-AMC using Pytorch and Python 3. We use AlexNet and VGG16 as our base models which we will compress. Since the receptive field of the first convolutional layer in AlexNet is too large $32\times32$ images, we reduced it to $3\times3$ when training on CIFAR-10. When training on Caltech256, we initialized the models with weights from models pretrained on ImageNet and tuned only the final classification layer. The accuracy and number of parameters of the base models are presented in Table \ref{tab:results_final_layer}.

\begin{table}[]
    \centering
    \begin{tabular}{c|c|c|c|c|c}
        Model & Dataset & Total ($\times 10^7$) & Dense ($\times 10^7$) & Conv ($\times 10^7$) & Acc \% \\ \hline
        
        \multirow{2}{*}{AlexNet} & CIFAR10 &  5.68 & 5.46 & 0.23 & 79.2\\
         & Caltech256 &  5.74 & 5.50 & 0.25 & 64.8\\\hline
        \multirow{2}{*}{VGG16} & CIFAR10 & 13.4 & 11.95 & 1.47 & 89.8\\
         & Caltech256 & 13.5 & 11.96 & 1.47 & 77.5 
    \end{tabular}
    \caption{\small The accuracy of the baseline models on CIFAR10 and Caltech256 and the number of parameters that they contain.}
    \label{tab:results_final_layer}
    \vspace{-10pt}
\end{table}

Since AMC does not define an order in which the layers must be shrunk, we must define one ourselves. We experiment with two orderings, namely, top down (TD) and round robin (RR). In TD we start from the penultimate layer, maximally shrink it and move down the network. In round robin (RR) we again start from the penultimate layer, but instead of maximally shrinking it we shrink each layer by at most a factor $\gamma$ and then move to the next layer.

We also introduce additional constraints in \texttt{LREShrink} (Algorithm \ref{alg:LRE-AMC}) to prevent the removal of neurons with independent observations and to stop removing neurons when $A$ becomes too error prone. Specifically, we do not apply the update if the average norm of the rows in the update is larger than the average norm of rows in the weight matrix i.e. ${\frac{1}{n^o_k}\sum_i\norm{\hat{W_{k+1}[i]} - W_{k+1}[i]}_2 > \frac{1}{n^o_l}\sum_i\norm{W_{k+1}}_2}$ or $\mathbbm{E}[|A_{.j}|\mathbbm{E}\left[Z^l_{.j}]\right] > 1$. To measure the effect of adjusting the network parameters using LRE, run experiments in which we do not adjust the network parameters using the LRE formalism presented in \ref{sec : readjust}. Instead, we prune the neurons with linearly dependent activations by simply dropping the corresponding columns from the weight matrix, and keeping the other columns as is.

Unless otherwise specified, we use the following hyperparameter settings. For experiments with AlexNet we use a learning rate of $10^{-4}$ and set $T=4$ in equation \ref{eq:amc-loss}. For experiments with VGG16 we use a learning rate of $5\times10^{-5}$ and set $T=5$. For both the models we set $\lambda=0.75$ in equation \ref{eq:amc-loss} and $\gamma=0.75$. During the fine-tuning phase, we tune the model for up to 50 epochs. We stop with the accuracy comes within $\epsilon=0.05$ of the precompression accuracy. If the accuracy on the held-out set does not improve for $3$ epochs we reduce the learning rate by 50\%. We stop tuning if the learning rate drops below $10^{-6}$.

\subsection{Results}
We present the percentage reduction in the number of model parameters, and the consequent loss in accuracy in Table \ref{tab:main-results}. The ``wAdj'' and ``noAdj'' settings correspond to the setting in which LRE is used and the setting in which LRE is not used. Under both these settings we demonstrate that our technique is able to decimate the number of parameters of AlexNet and VGG16, by pruning as much as 99\% of the model parameters.
\subsubsection{Top Down Shrinking}
We find that when we shrink the layers in top-down order we find that adjusting the model weights with LRE results in a significant reduction in model parameters. Adjusting the weights of AlexNet using LRE allows us to remove almost 30\% more parameters on CIFAR10 and 47\% more parameters on Caltech256, compared to when we did not adjust the weights. Furthermore, we observe that adjusting the weights allows us to prune additional neurons/filters from both, the dense and the convolution layers. This is an impressive result, not only because LRE-AMC able to reduce the number number of parameters in the network drastically but also because it yields better compression on the more difficult dataset. When we ran the same experiment with VGG16 we found that adjusting the weights using LRE results in slightly lower compression on CIFAR10 than however LRE is able to prune an additional 20\% of the model parameters, most of which are pruned from the dense layers.

\begin{table}[]
    \centering

\begin{tabular}{l|l|l|l|l|l||l|l|l|l}
\multirow{2}{*}{}& \multirow{2}{*}{Dataset} & \multicolumn{4}{l||}{AlexNet} & \multicolumn{4}{l}{VGG16}\\
& & $-\Delta_{A}$ & $-\Delta_{D}$ & $-\Delta_{C}$ & $-\Delta_{Acc}$ & $-\Delta_{A}$ & $-\Delta_{D}$ & $-\Delta_{C}$ & $-\Delta_{Acc}$ \\\hline
TD(noAdj) & CIFAR10 & 68.6 & 70.8 & 15.1 & \textbf{4.15} & 98.1 & 99.9 & 83.8 & 5.23\\
& Caltech256 & 37.3 & 38.8 & 03.8 & 6.35 & 46.5 & 51.9 & 02.1 & 5.44\\\hline
\multirow{2}{*}{TD(wAdj)} & CIFAR10 & 97.6 & \textbf{99.8} & 46.4 & 4.59 & 91.0 & 99.9 & 17.7 & 5.48 \\
& Caltech256 & 84.7 & 87.8 & 14.9 & 5.60 & 65.1 & 73.0 & 0.7 & 5.78\\\hline\hline
\multirow{2}{*}{RR(noAdj)} & CIFAR10 & \textbf{99.3} & 99.6 & \textbf{91.5} & 5.29 & 99.4 & 99.7 & 97.0 & 5.44\\
& Caltech256 & 75.8 & 75.9 & 74.2 & 5.33 & 80.9 & 82.0 & 72.0 & 5.87 \\\hline
\multirow{2}{*}{RR (wAdj)} & CIFAR10 & 97.3 & 97.5 & 91.2 & 5.36 & 97.4 & 97.5 & 96.2 & 5.34\\
& Caltech256 & 73.5 & 73.6 & 69.7 & 5.75 & 80.9 & 82.0 & 72.0 & 5.87                         
\end{tabular}
    \caption{\small The percentage reduction in the number of total parameters ($-\Delta_A$), dense layer parameters ($-\Delta_D$), convolutional layer parameters ($-\Delta_C$), and classification accuracy ($-\Delta_{Acc}$). }
    \label{tab:main-results}
    \vspace{-40pt}
\end{table}

\subsubsection{Round Robin Shrinking}
When we shrink the layers in a round robin fashion we find that we can achieve greater compression of the convolutional layers. Since the convolution layers scan the input, computing their activations involves a lot of floating-point operations (FLOPs). Reducing the number of convolutional filters greatly reduces the FLOPs of the model. Interestingly, performing round robin shrinking has a more significant impact on the total number of model parameter in AlexNet when the weights are not adjusted using LRE. In fact, under round robin shrinking not adjusting the weights yields \textit{slightly} better compression both in terms of reduction in the number of model parameters and the accuracy degredation. We also observe that under round robin shrinking, we achieve lower compression in terms of dense layer parameters on Caltech256 but we are able to prune away many more parameters from the convolutional layers. This seems to suggest that round robin shrinking would be ideal when minimizing FLOPs is more important than reducing memory consumption, while top-down shrinking should be preferred when memory consumption is to be optimized. 

\subsection{Analysis}
\subsubsection{Accuracy Error Tradeoff}
\begin{figure}
    \vspace{-10pt}
    \setlength{\abovecaptionskip}{-1.5pt}
    \setlength{\belowcaptionskip}{-14pt}
    \centering
    \subfloat[]{\includegraphics[width=0.3\textwidth]{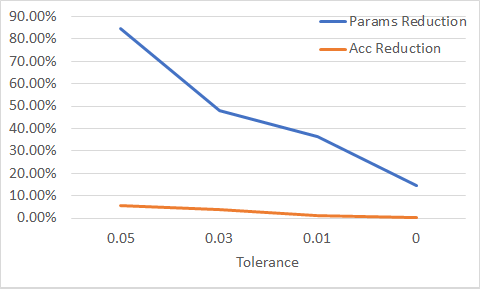}
        \label{fig:td-tolPlot}}
    \subfloat[]{\includegraphics[width=0.3\textwidth]{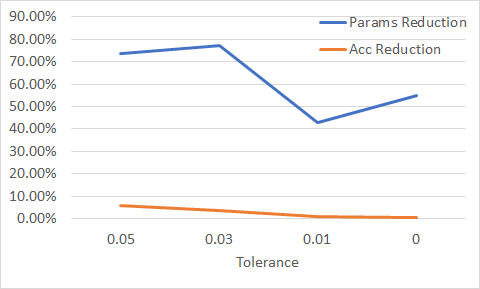}
        \label{fig:rr_tolPlot}}
    \caption{\label{fig:tolPlot}\small The change in the compression percentage of AlexNet as the accuracy tolerance is reduced from 5\% to 0\% (a) under top-down shrinking and (b) round robin shrinking  on Caltech-256. In both settings the weights are adjusted using LRE.}
\end{figure}

In this section we present experimental results that describe the compression-performance trade-off of our approach. As mentioned, we have used a tolerance of $\epsilon=0.05$ to limit the deterioration of accuracy during and after compression. In Figure \ref{fig:tolPlot} we plot the decrease in accuracy against the percentage of the parameters pruned for top-down and round robin shrinking of AlexNet on Caltech-256 at different values of $\epsilon$. Figure \ref{fig:td-tolPlot} exhibits the expected trend, in that, as we decrease $\epsilon$ both, the decrease in accuracy and the fraction of removed parameters decrease. 
We see that the the parameter reduction falls \textit{much} faster as we decrease $\epsilon$, indicating that that under the top-down shrinking scheme additional accuracy comes at a steep cost in compression performance. On the other hand, Figure \ref{fig:rr_tolPlot} exhibits a very different trend. As we decrease $\epsilon$ from 0.05 to 0.03 the compression improves, however, it deteriorates as when $\epsilon=0.01$ and improves again when $\epsilon=0.0$. Even though compression suffers when we set $\epsilon=0.0$, the deterioration is modest compared to the top-down shrinking. We do not have a reliable explanation for this phenomenon, because the repetitive nature of the round-robin shrinking approach makes its analysis complicated. 
It is entirely possible that removing neurons/filters in a certain order can lead to greater compression than removing neurons in some other order. The complexity arises if the optimal order spans across layers, something which the LRE framework does not account for. Though we do not prove it conclusively, the round robin shrinking approach seems to maintain compression even under very stringent accuracy constraints, and, therefore, shows promise as an effective model compression approach that could benefit from further study.

\subsubsection{Representation learning}

\begin{figure}[t!]
    \centering
    \begin{minipage}{0.32\textwidth}
        \includegraphics[width=1.\textwidth]{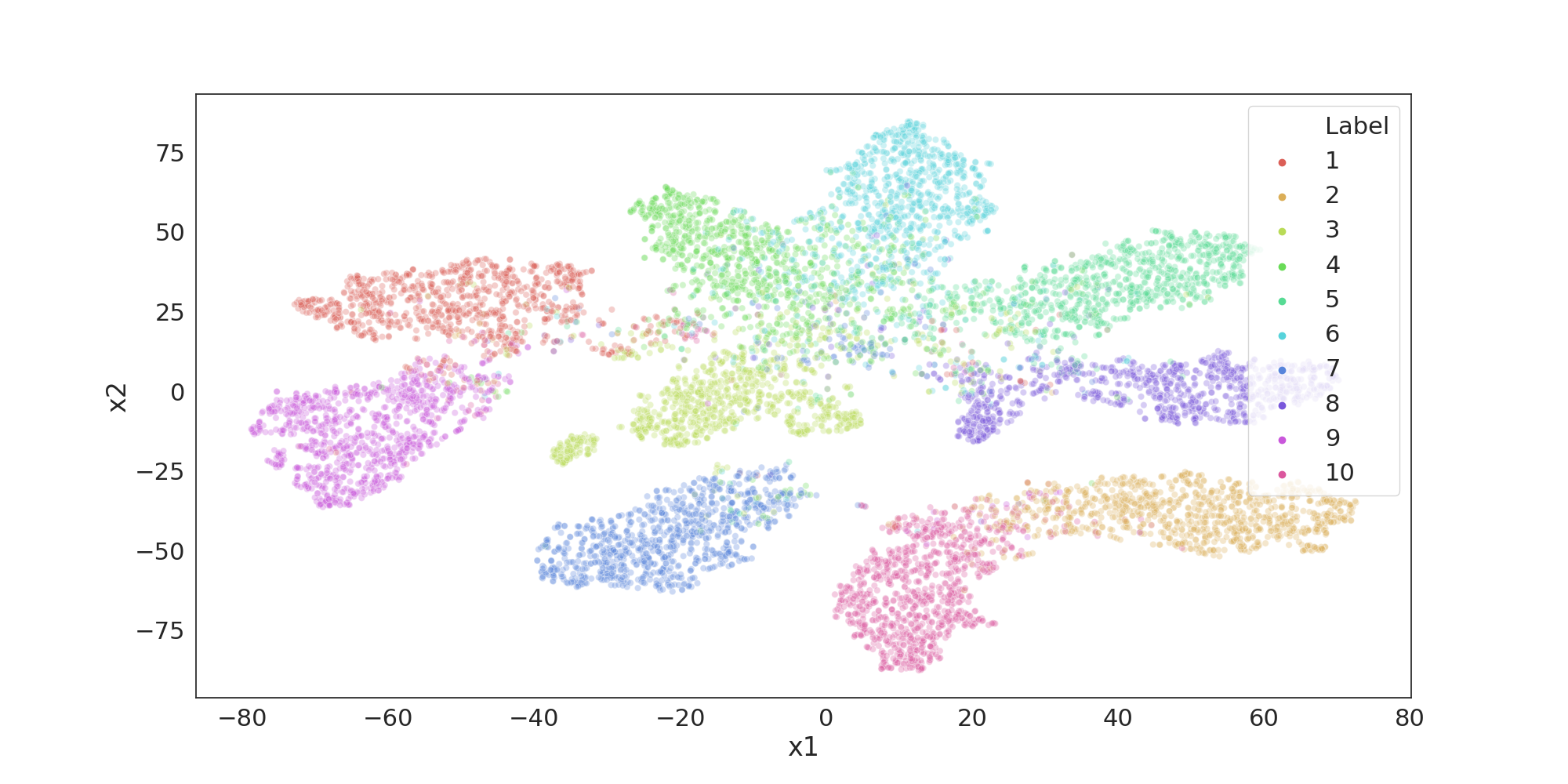}
        \includegraphics[width=1.\textwidth]{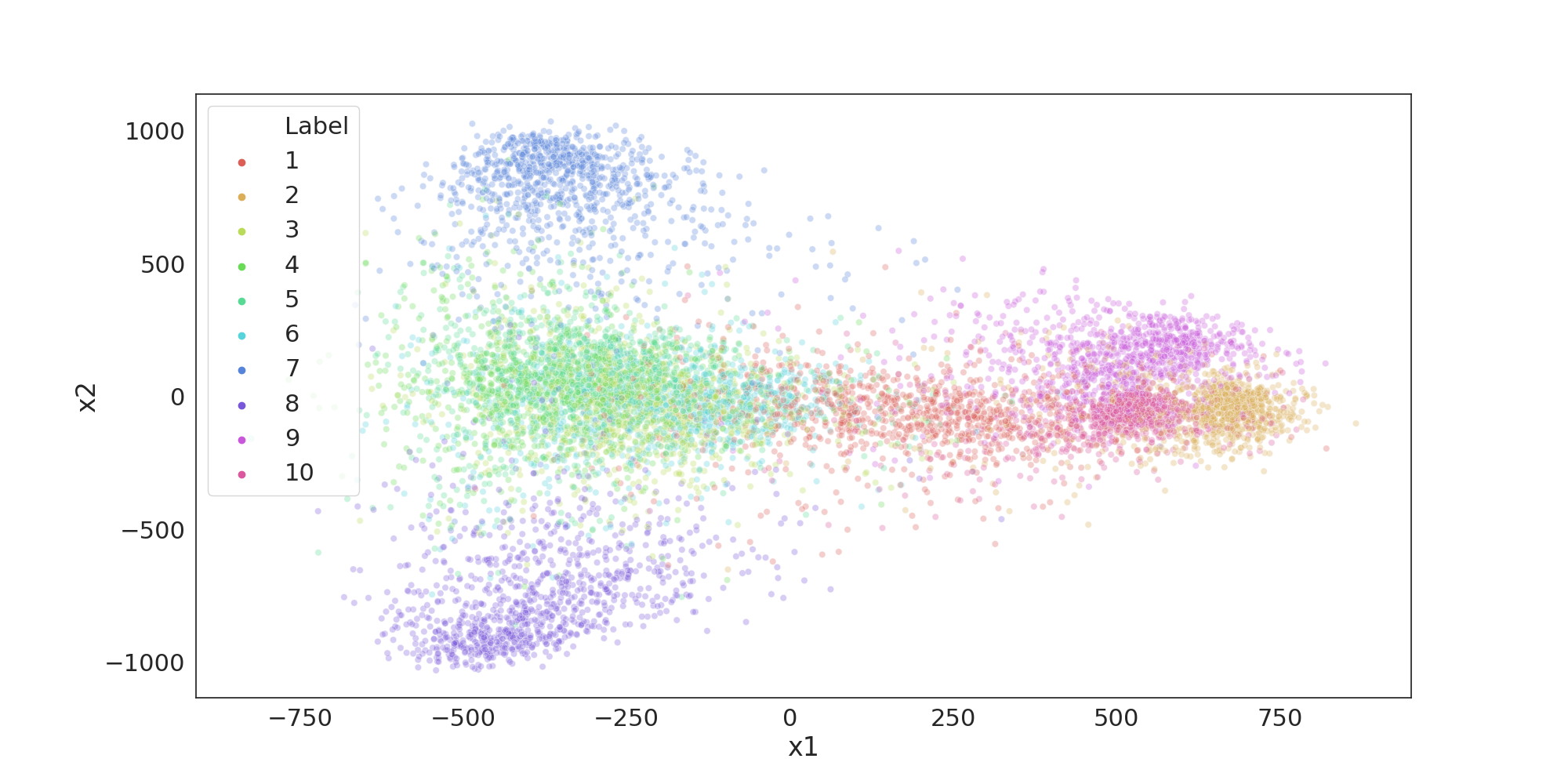}
    \end{minipage}
    \begin{minipage}{0.32\textwidth}
        \includegraphics[width=1.\textwidth]{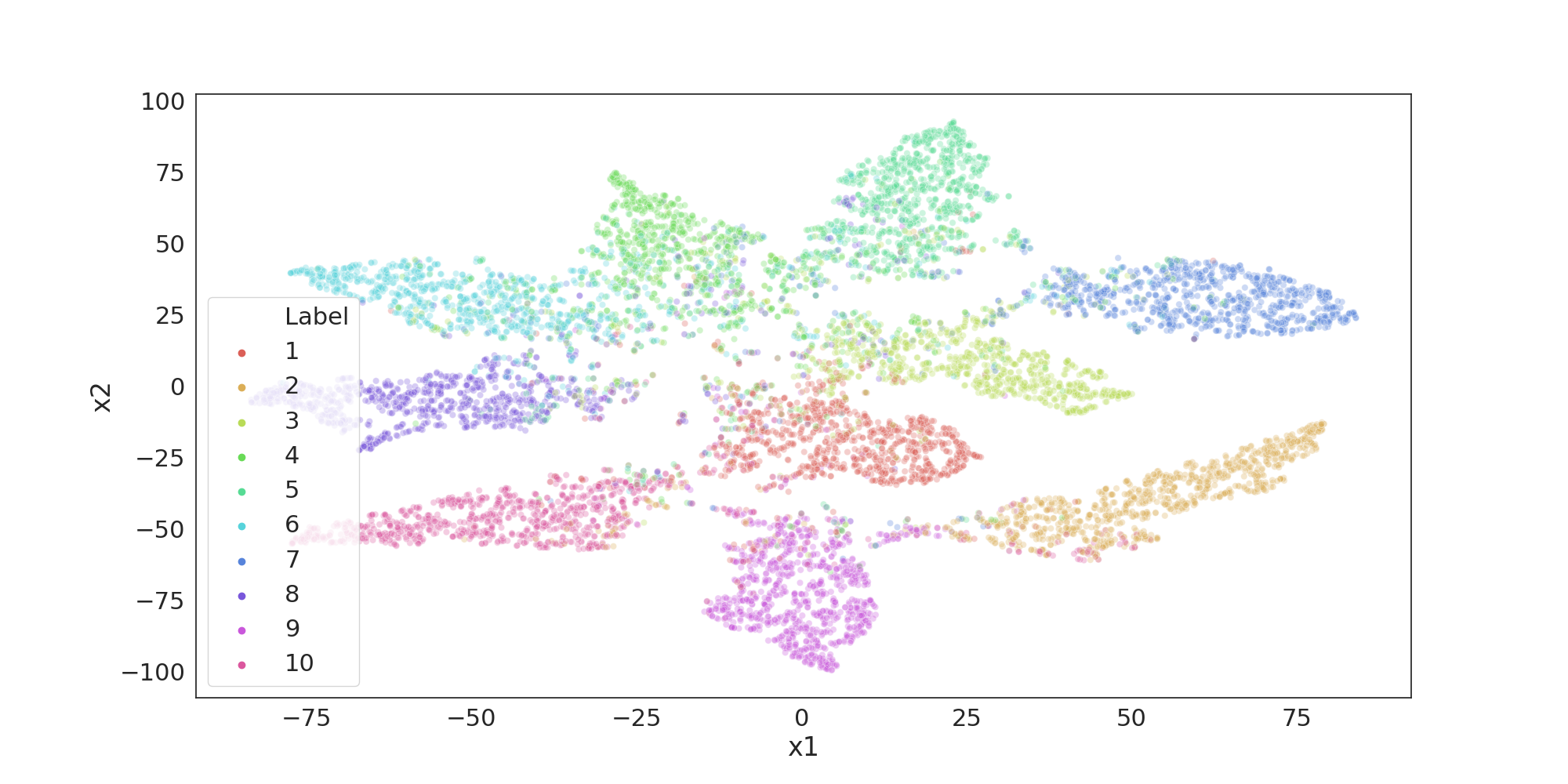}
        \includegraphics[width=1.\textwidth]{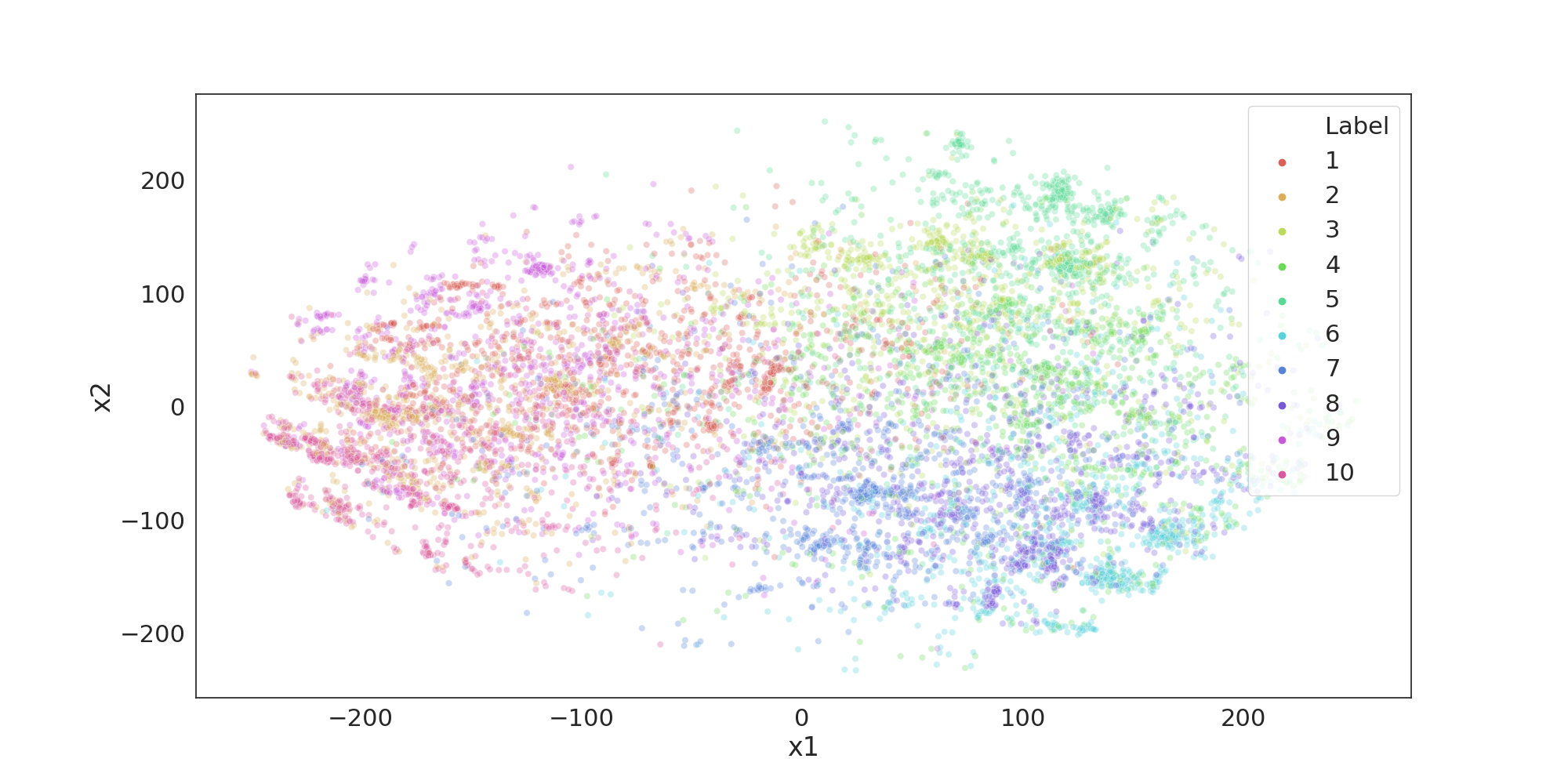}
    \end{minipage}
    \begin{minipage}{0.32\textwidth}
        \includegraphics[width=1.\textwidth]{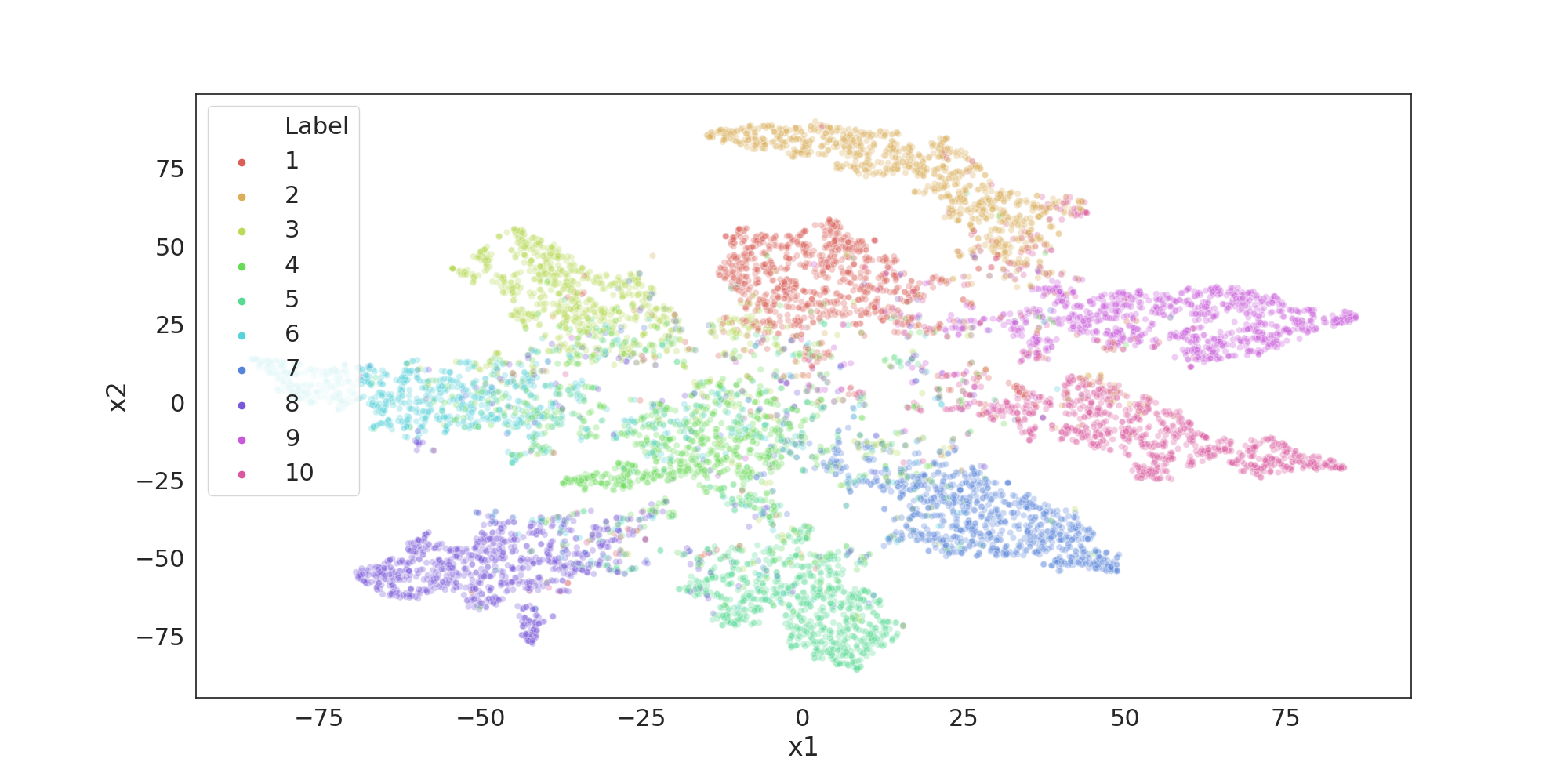}
        \includegraphics[width=1.\textwidth]{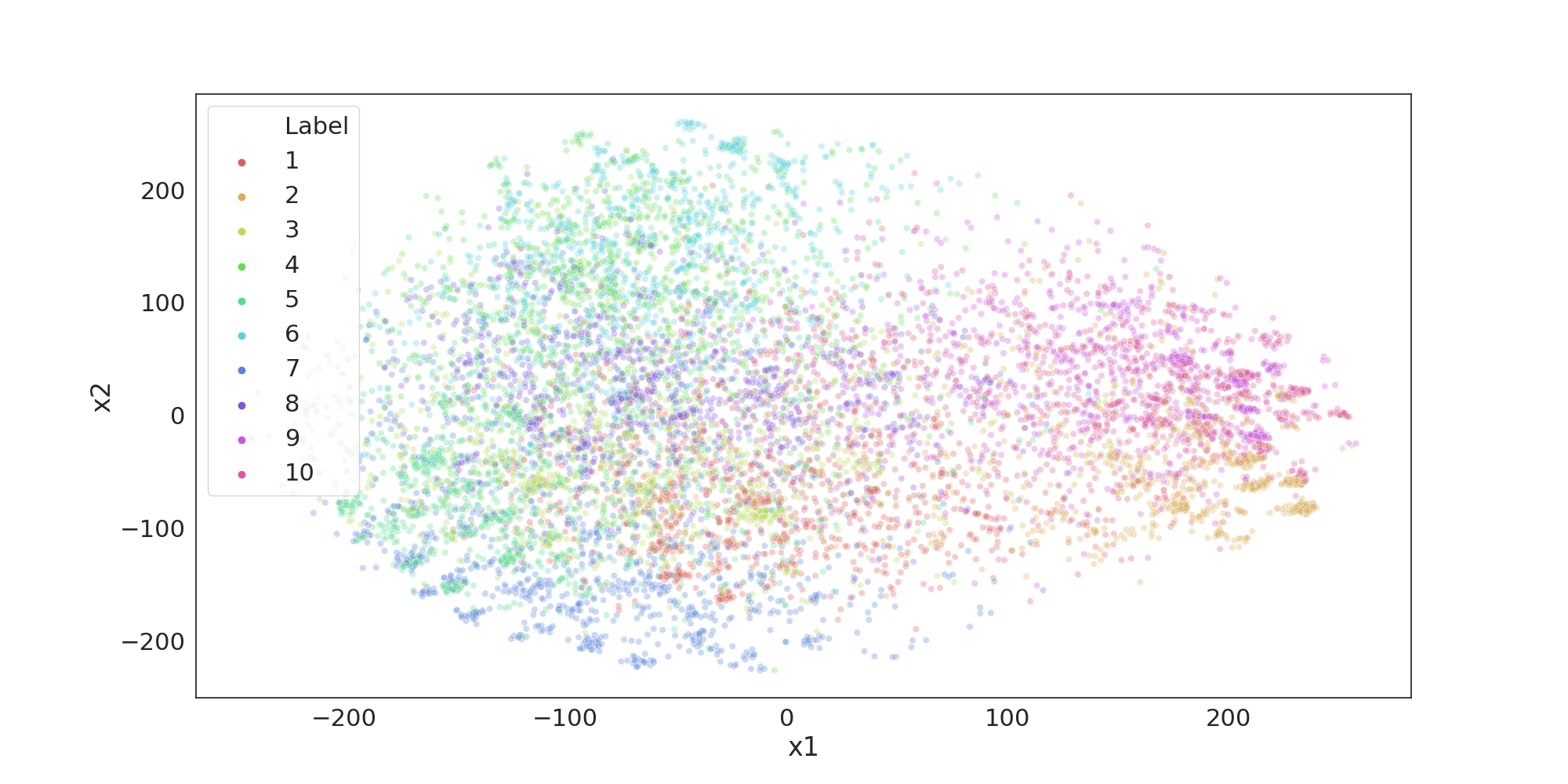} 
    \end{minipage}
    \caption{\label{fig : tsne}\small Low-dimensional representations of the final convolutional layer outputs of the VGG16 network trained on CIFAR. Points are test images colored by label. On the first line is t-SNE, on the second PCA. The left column corresponds to the original network ; the middle one to the compressed network without weight readjustment ; the right one to the compressed network with readjustment.}
    \vspace{-15pt}
\end{figure}

As we discussed earlier, one of the main use of a compressed representation on a task such as image recognition should be to provide useful pretrained representations for potential downstream tasks. When using performance-guided pruning, it is possible to degrade the learned representations while resorting on the final layers to use artifacts to maintain good performance on the recognition task. To make sure that our method isn't such a case, we provide some insight on the final convolutional layer of our VGG16 network trained on CIFAR through low-dimensional visualizations.
In \ref{fig : tsne} we plot t-SNE visualizations to observe separability. We can observe the compressed representations images are almost as separable as they are in the original network. Since t-SNE distorts the point clusters, we also plot PCA representations to assess the shape of the clusters. We can see that weight readjustment plays a significant role here : without it, clusters are more scattered and less convex, while with it they are more similar to the original network. This vouches for redundancy elimination and weight readjustment as compression methods that respect the semantics of the data, and that arguably are compatible with transfer learning between vision tasks. 

\section{Conclusion}
We have presented a novel neural model compression technique called LRE-AMC, which eliminates units (neurons or convolutional filters) whose activations are linearly dependent on the activations of other neurons in the same layer. Since entire units are pruned away, and not just individual parameters, the weight matrices in compressed model are smaller, which reduces the model's memory requirements and makes computation more efficient. We demonstrate the efficacy of LRE-AMC by applying it to AlexNet and VGG16 and show that we can remove more than 99\% of the units in both these models while suffering only a 5-6\% loss in accuracy on CIFAR-10. We have also applied LRE-AMC to the more difficult Caltech-256 dataset and achieved more than 80\% compression. Furthermore, we show that after compression the data remains separable in the model's intermediate layers, suggesting that the intermediate representation could carry sufficient information for transfer learning tasks. For future work we will explore methods of incorporating information from the derivatives of the subsequent layers to better estimate the effect of removing a unit on the overall output of the network, and prune neurons that minimally impact the output. We expect that this modification will result in smaller models and greater accuracy.

\bibliographystyle{splncs04}
\bibliography{refs}
\end{document}